\documentclass[letterpaper]{article} 
\usepackage{PRIMEarxiv}  
\usepackage[hyphens]{url}  
\usepackage{graphicx} 
\usepackage{natbib}  
\usepackage{caption} 
\usepackage{algorithm}
\usepackage{algorithmic}

\usepackage{newfloat}
\usepackage{listings}
\DeclareCaptionStyle{ruled}{labelfont=normalfont,labelsep=colon,strut=off} 
\floatstyle{ruled}
\newfloat{listing}{tb}{lst}{}
\floatname{listing}{Listing}

\usepackage{booktabs}

\usepackage{amsmath}
\usepackage{multirow}

\usepackage{xcolor}
\definecolor{selfrow}{HTML}{EFF6FF}
\definecolor{poscolor}{HTML}{DC2626}
\definecolor{negcolor}{HTML}{16A34A}

\colorlet{purple}{blue!40!red}

\colorlet{pink}{magenta!60!white}

\colorlet{pink}{magenta!60!white}

\title{Can We Trust LLM Judges: A Study of Capability-Dependent Biases and Multi-Judge Ensemble for Bias Calibration}
\author{ Gemma Zhang$^{1}$\thanks{Equal contribution.}, 
Prachi Badarayani$^{1}$\footnotemark[1], 
Asmi Kumar$^{1, 2}$\footnotemark[1], 
Sadid Hasan$^{1}$, 
Sulaiman Vesal$^{1}$ \\[1ex] $^{1}$Microsoft \\ $^{2}$Massachusetts Institute of Technology}

\begin{document}
\maketitle
\begin{abstract}

LLMs are increasingly used as automated judges for model training and evaluation, yet individual judges exhibit systematic biases that undermine reliability. Much of prior work has studied biases in pairwise LLM-as-a-judge settings; in this paper, we focus on absolute scoring tasks, which mirror more realistic use cases. Across four benchmarks and six models (36 judge-examinee pairs), we show that a model's task accuracy strongly predicts its judging accuracy (Pearson $r \geq 0.90$ on most models) and inversely predicts its directional bias ($r \leq -0.83$), but that accuracy alone does not ensure fair evaluation: more capable examinee models consistently receive more lenient judgments from all judges ($r \geq 0.83$). To address this, we propose calibrated weighted majority voting (WMV), an ensemble evaluation method that aggregates multiple LLM judges weighted by online estimates of their false-positive and false-negative rates. We introduce a disagreement-based estimator that derives these error rates purely from inter-judge agreement patterns, requiring no ground-truth labels or task metadata. In a simulated experiment with shifting task distributions, our label-free WMV tracks an oracle with perfect error-rate knowledge to within 0.5 percentage points on average, outperforming both individual judges and unweighted majority voting. These results demonstrate that principled multi-judge calibration can simultaneously improve accuracy and correct for systematic leniency without requiring labeled data, offering a scalable path to reliable automated evaluation as model capabilities increase.
\end{abstract}

\section{Introduction}
Large language models (LLMs) are increasingly used as judges of other models' outputs, often called LLM-as-a-judge \cite{zheng2023judging}. These judgments now power benchmarks, model selection, and reward modeling~\cite{zheng2023judging, lambert2025rewardbench, mazeika2024harmbench}. LLM-as-a-judge is appealing because it offers a scalable and intuitive alternative to human evaluation, and recent work shows that strong models can approximate human pairwise preferences in open-ended evaluation settings \citep{zheng2023judging, chiang2024chatbot, dubois2023alpacafarm}. At the same time, the judge is itself a model, so its decisions can reflect model-specific biases that are separate from response quality.

Prior work has identified several such biases, showing that LLM judges are sensitive to details of the evaluation setup. Judges may prefer responses based on their position in the prompt, reward longer answers regardless of correctness, favor particular writing styles, or agree with user-suggested conclusions. They may also show self-preference, favoring outputs from their own model family \cite{zheng2023judging, chen2024humans, panickssery2024_240413076, wataoka2024_241021819}. These findings raise concerns about whether LLM-as-a-judge measures answer quality or also rewards features unrelated to the task.

\begin{figure}
    \centering
    \includegraphics[width=0.5\linewidth]{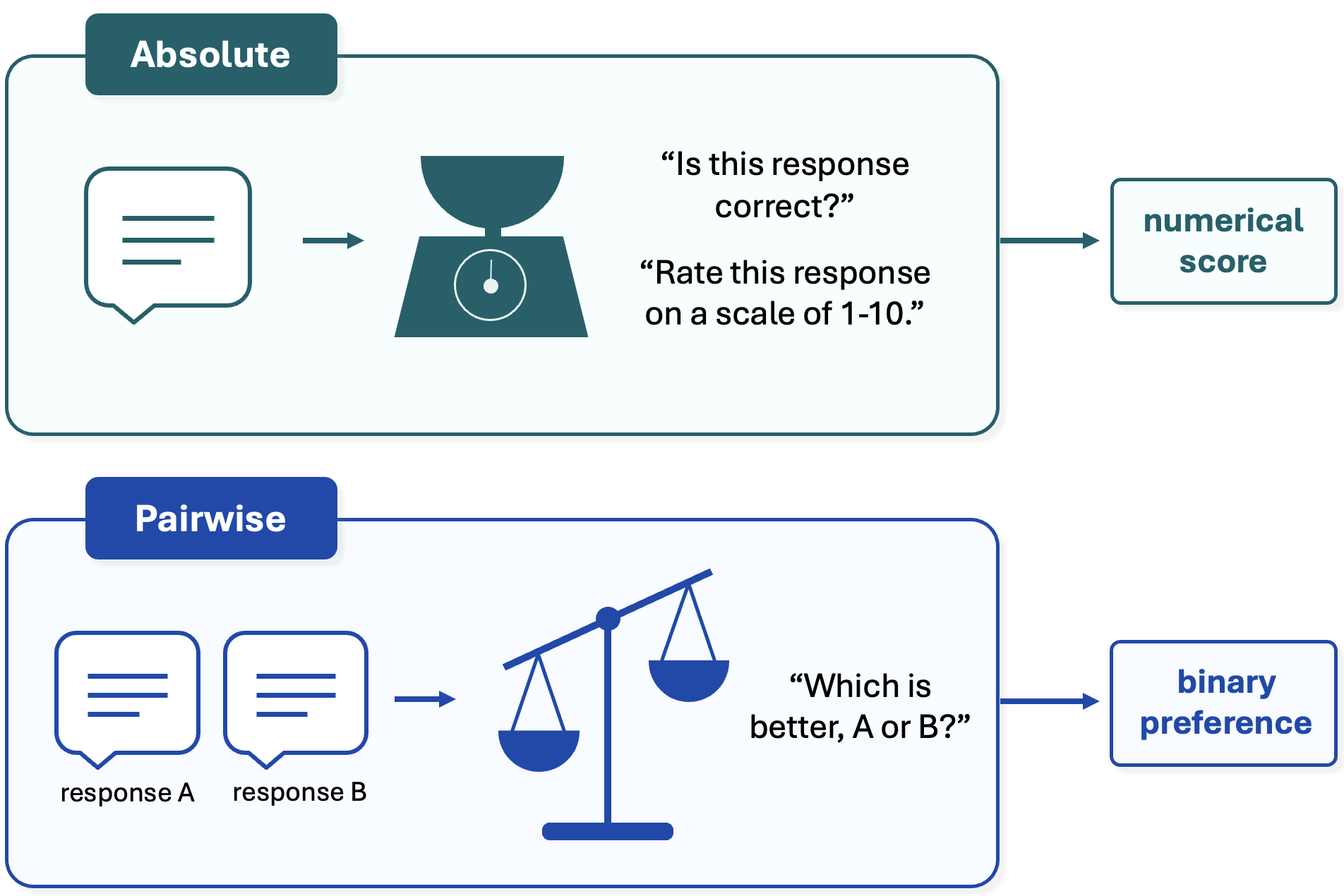}
    \caption{Absolute scoring assigns an independent quality score to a single response (e.g., marking it as correct or incorrect), whereas pairwise scoring compares two responses and selects the better one.}
    \label{fig:eval_compared}
\end{figure}

Much of this work, however, studies pairwise preference settings, where the judge compares two responses and selects the better one (Figure \ref{fig:eval_compared}). Many practical evaluation workflows instead require an absolute decision about a single response (rubric-based scoring, pass/fail test suites, safety gatekeeping), where the question is not which response is better but whether a given response is correct, safe, or acceptable. The main failures in this setting are false positives, where incorrect answers are accepted, and false negatives, where correct answers are rejected. These errors can silently distort model rankings, corrupt training signals, and influence deployment decisions, yet they are invisible without ground-truth labels.

This raises a central question for LLM-as-a-judge evaluation: how does a judge's behavior depend on the identity and capability of the model being evaluated? Do judges treat competing models symmetrically, or do their errors systematically favor some models over others? While recent work studies harmful self-preference in pairwise objective tasks~\cite{chen2025_250403846}, absolute scoring exposes a different failure mode: leniency toward particular examinee models, including models other than the judge itself. We study how LLM judges distribute their errors across models in this setting, since this is central to assessing whether they are reliable enough to stand in for human evaluators. We find that the answer is nuanced: judges systematically favor more capable models' outputs, a pattern that is broader than self-preference and intensifies with model capability.

Beyond characterizing these biases, we ask whether they can be corrected without sacrificing scalability. We propose \emph{calibrated weighted majority voting} (WMV), which weighs each judge's vote by an online estimate of its false-positive and false-negative rates derived purely from inter-judge agreement patterns, requiring no ground-truth labels. Our contributions are:
\begin{enumerate}
    \item A systematic study of absolute-scoring bias across four objective benchmarks (math, code, reading comprehension), revealing that stronger examinee models receive systematically more false-positive leniency from \emph{all} judges, not just from themselves.
    \item Calibrated weighted majority voting for bias mitigation with three online FPR/FNR estimators (Bayesian, stratified, and
    disagreement-based), the last of which requires no ground-truth labels.
    \item A temporal-drift evaluation showing that label-free WMV matches an oracle with perfect error-rate knowledge, outperforming both individual judges and unweighted majority vote under non-stationary task distributions.
\end{enumerate}

\section{Related Work}
LLM-as-a-judge has become a common evaluation paradigm, but prior work has shown that judges can exhibit systematic biases unrelated to response quality. These include position bias~\cite{zheng2023judging, shi2025_240607791}, verbosity bias~\cite{saito2023_231010076}, style and formatting bias~\cite{stureborg2024_240501724}, self-preference~\cite{zheng2023judging, panickssery2024_240413076}, and agreeableness bias~\cite{jain2025_251011822}. Closely related to our focus, \cite{wataoka2024_241021819} links judge preference to perplexity, suggesting that outputs from more capable models may receive higher evaluations independent of correctness. In addition, \cite{chen2025_250403846} study self-preference on objective benchmarks in a pairwise setting and analyze legitimate vs harmful self-preference.

Several works have proposed methods for measuring or mitigating these effects. \cite{ye2024_241002736} introduce CALM, a taxonomy and automated framework for bias detection, while \cite{chen2025surfacemeasuringselfpreferencellm} propose the DBG score to disentangle self-preference from true quality differences. On the mitigation side, style perturbations can reduce self-preference but may not fully remove self-recognition~\cite{mahbub2025_251205379}, and reasoning-based correction can reduce biased evaluations~\cite{yang2026_250517100}. Most closely related, \cite{chen2025_250403846} study harmful self-preference on verifiable tasks and show that chain-of-thought can reduce it in pairwise comparisons.

Our work differs from prior studies in two key ways. First, we move from pairwise comparisons to absolute scoring, the setting used in most rubric-based evaluation pipelines, reward model training, and automated test suites, where biases manifest as asymmetric error rates rather than preference orderings. Second, we go beyond diagnosis: we propose and evaluate an unsupervised mitigation that corrects these biases online without ground-truth labels.

\section{Methodology}
\subsection{Datasets}
We evaluate LLM judge bias on four objective benchmarks spanning reading comprehension, math reasoning, and code generation: QuALITY~\citep{pang2022quality}, GSM8K~\citep{cobbe2021gsm8k}, MBPP~\citep{austin2021mbpp}, and AIME~\citep{matharena2025aime}. These datasets provide unambiguous ground-truth answers, allowing us to measure false positives and false negatives directly rather than relying on subjective preference judgments.

We use the test split for GSM8K (1{,}319 problems) and MBPP (500 problems), the train split for AIME~2025 (30 problems; the only available split), and the dev split of QuALITY. Further details are provided in Appendix~\ref{app:experimental-details}. Together, these benchmarks cover multiple reasoning domains and difficulty levels, with GSM8K and AIME representing easier and harder mathematical reasoning settings, respectively.


\subsection{Metrics} 
\label{subsec:metrics}
We use error-rate metrics rather than aggregate accuracy alone, since accuracy can mask systematic biases when a judge is more lenient toward some examinees than others.
\begin{enumerate}
    \item \textbf{False positive rate (FPR, leniency)} measures how likely a model is to rate an incorrect answer as correct, and it is defined as: $\frac{\# FP}{\# FP + \# TN}$
    \item \textbf{False negative rate (FNR, strictness)} measures how likely a model is to rate a correct answer as incorrect, and it is defined as: $\frac{\# FN}{\# FN + \# TP}$ 
    \item \textbf{Directional bias (leniency - strictness)} measures the tendency of a model to erroneously accept wrong answers rather than erroneously reject correct ones. A neutral judge should have 0 directional bias.
    \item \textbf{Judge accuracy}: $\frac{\# TP + \# TN}{\mathrm{Total}}$
    \item \textbf{$\Delta$FPR and $\Delta$FNR}: The difference in FPR (or FNR) between a judge's evaluation of its own model's responses versus another model's responses. A positive $\Delta$FPR indicates the judge is more lenient toward its own (or a particular model's) wrong answers.
\end{enumerate}
\subsection{Models}
We evaluate six models spanning three families: GPT-5~\citep{openai2026gpt5},
GPT-4.1-mini~\citep{openai2025gpt41}, and o4-mini~\citep{openai2025o3o4mini}; Llama-3.3-70B-Instruct~\citep{grattafiori2024llama3} and
Llama-4-Maverick~\citep{meta2025llama4}; and
DeepSeek-V3.2~\citep{deepseek2025v32}. This enables us to distinguish self-preference from within-family preference and cross-family model preference. Each of the judge-examinee combinations is evaluated across all benchmark questions.

\paragraph{Judge input format.}
For each task, the judge model receives the examinee's full response (not just the final answer) in a single-turn prompt and outputs ``correct'' or ``incorrect.'' For GSM8K and AIME, the judge sees ``Final Answer: $\langle$answer$\rangle$\textbackslash n\textbackslash nReasoning: $\langle$solution trace$\rangle$,'' which includes the model's complete chain of reasoning. For MBPP, the judge receives the submitted Python code along with the problem statement and test cases. For QuALITY, the judge sees the full article passage, the question, and the model's answer choice together with its justification. This means that length and stylistic properties of the reasoning trace are visible to the judge in all tasks, motivating the length confound analysis in Section~\ref{sec:results:length}.

All models are accessed via Azure API endpoints with near-deterministic decoding ($t{=}0$) where supported; reasoning models (o4-mini, GPT-5) use the API-enforced $t{=}1$. Full experimental details including language model hyperparameters, API configuration, dataset splits, bootstrap parameters, and simulation settings are provided in Appendix~\ref{app:experimental-details}.

\section{Experiments and Results}
To evaluate LLM judge biases and make practical recommendations, we set out to answer the following questions: 
\begin{enumerate}
    \item Are stronger models more accurate and less biased judges of others? 
    \item Does higher judging accuracy imply less biased judging? 
    \item Do more accurate models get more leniency from judge LLMs? 
\end{enumerate}
Based on our findings in answering these questions, we will make practical recommendations for bias mitigation in judge LLMs. 

\begin{figure}[t]
    \centering
    \includegraphics[width=0.7\linewidth]{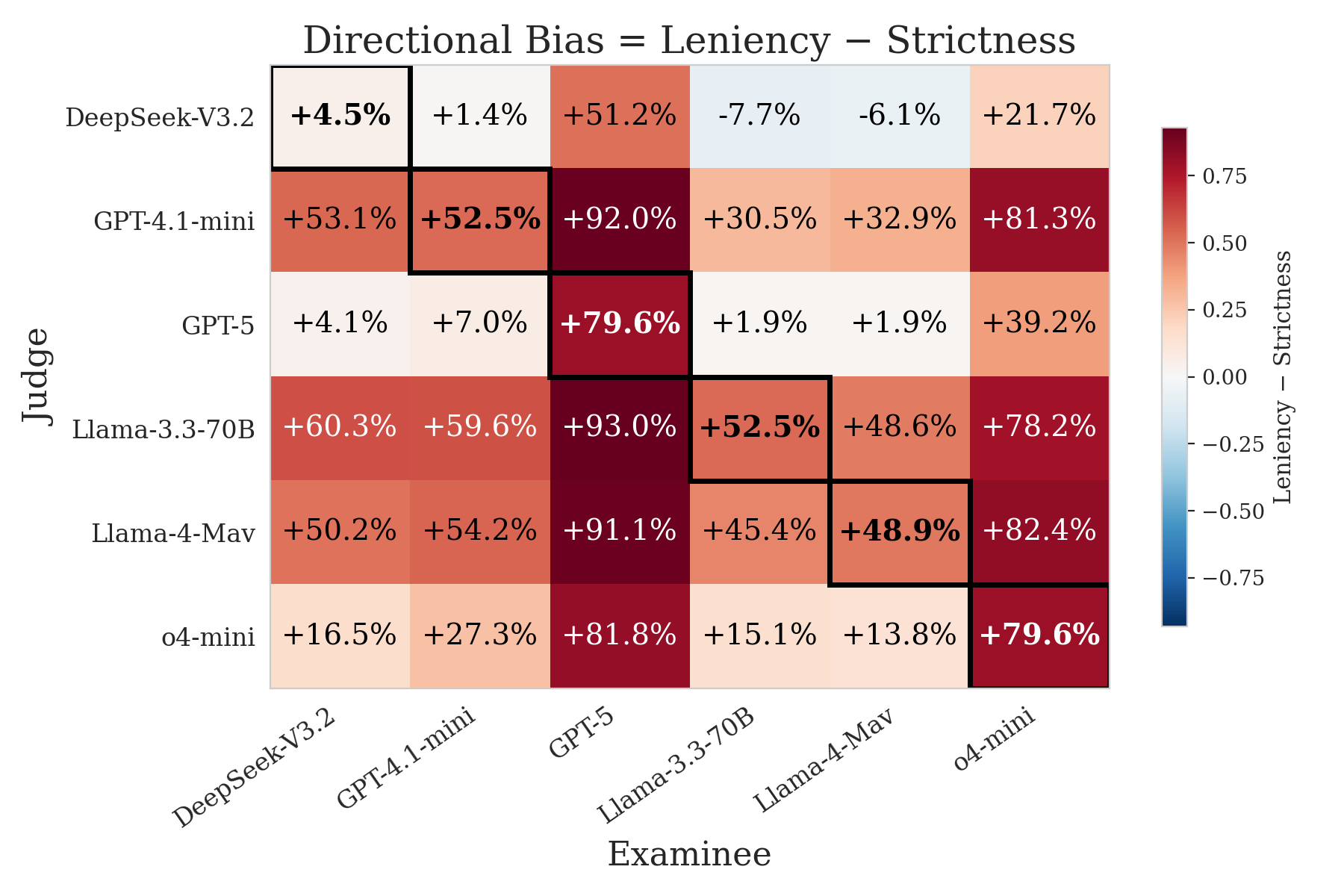}
    \caption{Directional bias for each judge-examinee model pair on QuALITY dataset; diagonal cells denote self-evaluation and off-diagonal cells denote cross-model judging.}
    \label{fig:directional_bias}
\end{figure}

Figure~\ref{fig:directional_bias} presents an example full directional bias matrix across all judge-examinee pairs, aggregated over the QuALITY benchmark. Each cell shows
$\mathrm{DirectionalBias}(J,E) = \mathrm{FPR}(J,E) - \mathrm{FNR}(J,E)$
together with the number of evaluated instances~$n$. Green-bordered cells on the diagonal mark self-evaluation (judge = examinee). Red indicates net leniency; blue indicates net strictness; white indicates a balanced judge. 

For clarity and brevity, we only present results for the QuALITY dataset in Figure \ref{fig:directional_bias} but note that other datasets generally follow the trends that we observe. We include full results of all experiments across all judge-examinee pairs in Section~\ref{appx:correlations}. 

\subsection{More accurate models are better judges}
\label{subsec:acc_vs_judge_acc}
There are two dimensions to judging quality: 1) overall judging accuracy; 2) preference bias in judging. To look at these two facets, we plot the accuracy and directional bias of a judge model as a function of its own accuracy on the QuALITY dataset in Figure~\ref{fig:correlations}(A) and Figure~\ref{fig:correlations}(B), respectively. 

In Figure~\ref{fig:correlations}(A), we look at whether the accuracy of a model on a task translates to how accurately it acts as a judge, on a set of responses from a given examinee model on the same task. As shown in the figure, in most cases, there is a strong positive correlation between a model's own response accuracy and its accuracy judging responses of other models, in agreement with findings from~\cite{chen2025_250403846, mahbub2025_251205379}. We measure the strength of the correlation using Pearson's correlation coefficient $r = \frac{\mathrm{Cov}(X, Y)}{\sigma_X \sigma_Y}$, where $\sigma_{X}$ and $\sigma_{Y}$ denote standard deviations. 

Here, we observe that when GPT-5 is the examinee model, the correlations are much weaker (Table~\ref{tab:corr-panel-a}). We attribute this to two factors. First, GPT-5's high task accuracy (93--97\% across benchmarks) means it produces very few wrong answers, so each judge's FPR on GPT-5 is estimated from a small denominator and is therefore noisy. Second, the wrong answers GPT-5 \emph{does} produce tend to contain elaborate, well-structured reasoning that is difficult for any judge to catch, regardless of that judge's own capability. In this regime, all judges converge toward similarly high FPR on GPT-5's errors, flattening the relationship between judge accuracy and judge performance on this examinee. This foreshadows the capability-dependent leniency we analyze in Section~\ref{subsec:examinee_acc_vs_bias}.


In Figure~\ref{fig:correlations}(B), instead of a model's accuracy as a judge, we look at its preference bias given to other models, as mediated by the directional bias metric. We first note that the DeepSeek model is overall stricter as a judge than other models (i.e., smaller directional bias values), as seen in Figure~\ref{fig:directional_bias}. In most cases, there is a strong negative correlation between a model's accuracy on a task and its directional bias when evaluating other models on the same task, especially when we exclude Deepseek's data points. With the Deepseek judge data points excluded from linear fitting, the Pearson's absolute value of $r$ on the average across all examinees is $>0.9$ and indicates very strong correlation (see Appendix~\ref{appx:correlations}). In other words, the more accurate a model is, the more neutral it is when judging other models.

As an aside, we note that the models almost universally exhibit positive directional bias, which indicates strong LLM judge agreeableness bias. 

\subsection{More accurate judges are generally less biased}
\begin{figure}[h]
\centering
    \includegraphics[width=0.7\linewidth]{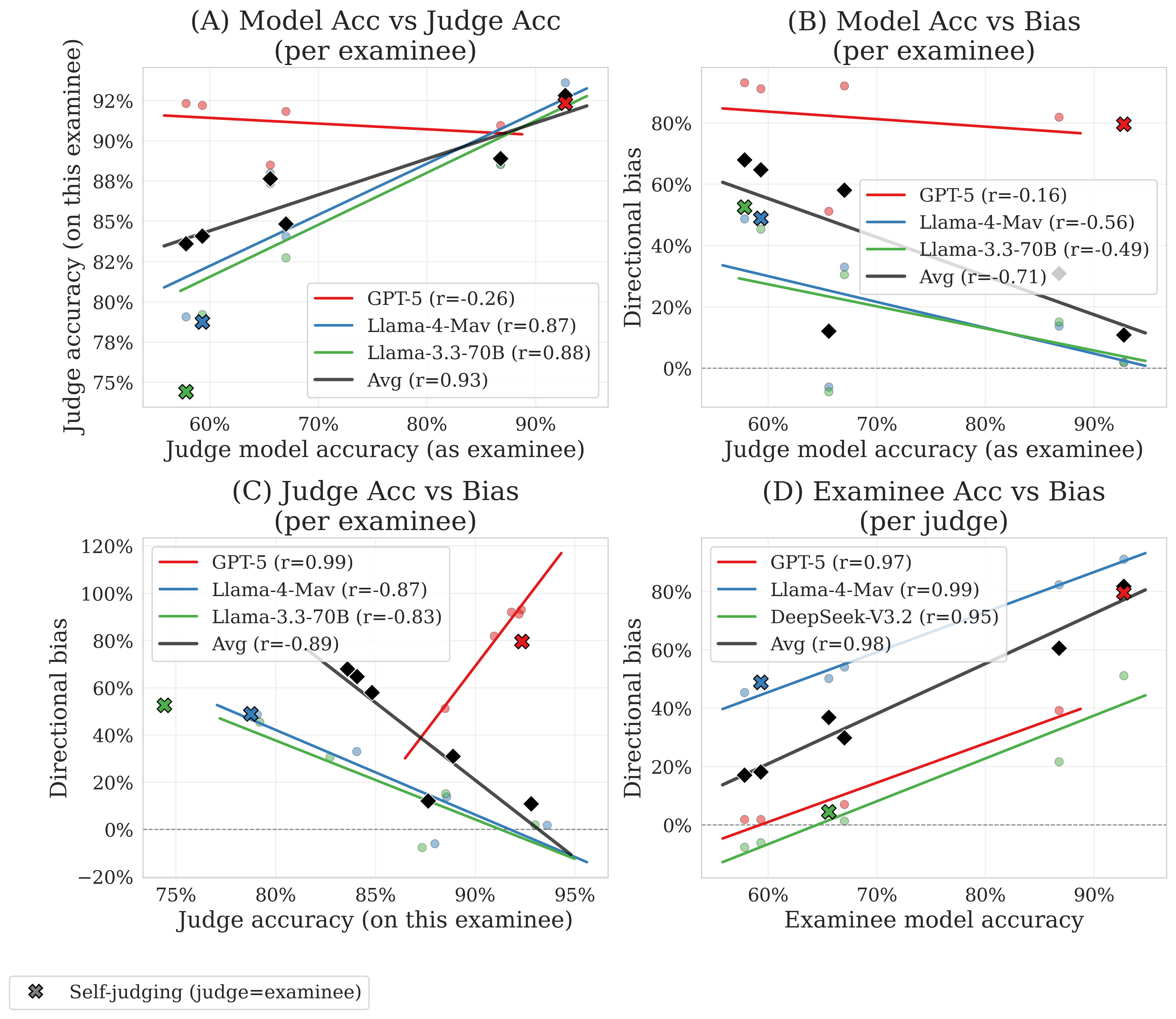}
    \caption{Correlations between model accuracy and judging behavior on QuALITY (reading comprehension), shown for three models and the average over all models. (A) Judge model accuracy as an examinee vs. its accuracy when judging each examinee. (B) Judge model accuracy as an examinee vs. directional bias toward each examinee. (C) Judge accuracy on a specific examinee vs. directional bias toward that examinee. (D) Examinee model accuracy vs. directional bias received from each judge. Regression lines are fit on cross-judging pairs only (excluding self-judging). The black line with diamond markers shows the averaged relationship across all models (one point per model, averaged over all its cross-judging counterparts). X markers indicate self-judging pairs where judge and examinee are the same model. Pearson's coefficients are shown in parentheses.}
    \label{fig:correlations}
\end{figure}

In Figure~\ref{fig:correlations}(C), we plot the directional bias as a function of judge accuracy on each examinee model. We note again that DeepSeek stands out as an outlier as a judge. On most examinee model responses, we see a strong negative correlation between judge accuracy and judge bias, especially when we disregard data points for which DeepSeek is the judge model. In other words, more accurate judges are generally less biased. Note that because correlations are not transitive, this is not a trivial result that can be deduced from the conclusions of Section~\ref{subsec:acc_vs_judge_acc}.

However, we note the strongly positive correlation between judge accuracy and bias when GPT-5 is the examinee. In addition, we also see that most models, when judging GPT-5, incur a high directional bias indicating a high degree of leniency. This may be due to GPT being a stronger model in this task compared to other models we ran experiments with, as shown in Table~\ref{tab:model_accuracy}. This can also be seen in the plot in which o4-mini is the examinee model: the models with lower accuracy acting as judges leads to higher directional bias, while GPT-5 judging o4-mini has comparatively lower leniency as it has a higher accuracy on the QuALITY dataset than o4-mini. 

This observation leads to the question of whether a stronger (more accurate) model generally gets preferentially lenient treatment by judge models, which we will discuss in Section~\ref{subsec:examinee_acc_vs_bias}. 

\subsection{More accurate models are more leniently judged}
\label{subsec:examinee_acc_vs_bias}
To look at whether more accurate models get more leniency when it is judged by other models, in Figure~\ref{fig:correlations}(D), we plot the directional bias as a function of examinee model accuracy, for each judge model. The strong positive correlation in each case indicates that stronger models are judged more leniently by LLM judges. This indicates capability-dependent leniency: judges are more likely to accept wrong answers from stronger examinee models, even when the judge and examinee are from different model families. Our finding highlights the need for practitioners to pay careful attention to bias mitigation as model capability continues to improve rapidly.

We note that self evaluation datasets generally fall closely along the linear regression lines (which are plotted without self-evaluations). This indicates that elite model preference significantly dominates over self-preference for most models we tested. The only exception in our studies is GPT-5, whose self-preference bias can be clearly observed in Figure~\ref{fig:correlations}(D).
Appendix~\ref{subsec:add_exp} confirms this with 95\% bootstrap CIs on $\Delta$FPR (Table~\ref{tab:deltas}) and a harmful-subset analysis (Table~\ref{tab:harmful}).




\begin{table}[h]
\centering
\small
\caption{Model accuracy (\%) on each evaluation task.}
\label{tab:model_accuracy}
\begin{tabular}{lcccc}
\toprule
Model & QuALITY & MBPP & GSM8K & AIME \\
\midrule
DeepSeek-V3.2  & 65.6 & 90.6 & 63.3 & 53.1 \\
GPT-4.1-mini   & 67.0 & 85.0 & 45.6 & 54.6 \\
GPT-5          & 92.7 & 96.6 & 95.9 & 95.0 \\
Llama-3.3-70B  & 57.8 & 75.8 & 35.1 &  3.4 \\
Llama-4-Maverick & 59.3 & 80.4 & 55.9 & 27.2 \\
o4-mini        & 86.8 & 95.0 & 95.3 & 91.7 \\
\bottomrule
\end{tabular}
\end{table}

\subsection{Summary} 
In light of these results, we make the following recommendations for practitioners to mitigate preference bias in LLM judges: 
\begin{enumerate}
    \item To judge a task, use the model with the best response accuracy on this task. 
    \item If a golden evaluation set is available, to minimize leniency bias, use the model that has the highest \textit{judging} accuracy on this golden set.
    \item As model capabilities increase, judging them impartially becomes more challenging, which leads us to propose a mitigation method, discussed in Section~\ref{sec:mitigation}. 
\end{enumerate}

\section{Calibrated Ensemble Evaluation as Mitigation}
\label{sec:mitigation}
When we know the judge accuracy on a given task, selecting the judge with highest accuracy produces lowest directional bias. However, online systems may encounter temporally changing task type and difficulty~\cite{jain2024livecodebench, white2024livebench}, thus making it hard to assign a single best judge for evaluation. Given the magnitude of capability-dependent leniency documented above, we evaluate cross-judge ensembles as a practical mitigation. 

\textbf{Static majority voting.}
We exclude the self-judge and aggregate the remaining cross-judges via simple majority vote (MV).
Table~\ref{tab:mv} reports accuracy and leniency (FPR) averaged across all examinee models.

\begin{table}[t]
\centering
\caption{Cross-judge majority voting vs.\ self-judging. Averages across all examinee models per task.}
\label{tab:mv}
\small
\begin{tabular}{lrrrr}
\toprule
& \multicolumn{2}{c}{\textbf{Accuracy\%}} & \multicolumn{2}{c}{\textbf{Leniency (FPR)\%}} \\
\cmidrule(lr){2-3}\cmidrule(lr){4-5}
\textbf{Task} & \textbf{Self} & \textbf{MV} & \textbf{Self} & \textbf{MV} \\
\midrule
GSM8K   & 89.3 & 96.8 & 39.1 & 20.7 \\
AIME    & 74.3 & 88.6 & 60.3 & 44.9 \\
MBPP    & 84.2 & 90.9 & 56.3 & 33.9 \\
QuALITY & 77.2 & 82.4 & 59.4 & 41.9 \\

\bottomrule
\end{tabular}
\end{table}

MV improves accuracy by 5--14 percentage points and reduces leniency by 15--22 percentage points across all three tasks.
However, MV is not a panacea: when the \emph{strongest} model judges its own outputs (e.g., o4-mini on GSM8K), the weaker cross-judges can be outvoted by their own errors; on GSM8K, MV achieves 94.4\% accuracy vs.\ 97.2\% for the self-judge, because most cross-judges are too lenient on the few items the examinee got wrong.
The practical challenge is that the identity of the ``best judge'' varies by task and is unknown without ground truth.

\textbf{Temporal drift and online calibration.}
In practice, the distribution of evaluation problems changes over time: coding contest platforms introduce new problems with shifting difficulty~\cite{jain2024livecodebench}, and user queries to chat systems evolve as new models are released~\cite{chiang2024chatbot}.
Our own results confirm that no single judge is uniformly best across tasks (Table~\ref{tab:model_accuracy}), so static judge selection is insufficient when the task mix is non-stationary.
The question, then, is whether an online estimator can track shifting judge reliability without labeled data.

To test this, we simulate a 12-week deployment that mimics a realistic difficulty shift: the task mix drifts linearly from 85\% GSM8K (easy) / 15\% AIME (hard) to the reverse, sampling 40 items per week. This design stresses the estimators because each judge's effective FPR changes as harder problems become more prevalent; an estimator that cannot adapt will increasingly misweight judges.

\textbf{Online weighted majority vote.} Each judge $j$ has task-specific error rates $\mathrm{FPR}_j$ and $\mathrm{FNR}_j$ that are unknown and must be estimated online.
The ensemble aggregates judges via \emph{weighted majority vote} (WMV): for an item with verdicts $\{v_j\}_{j \in \mathcal{J}}$, the weighted ensemble predicts ``correct'' if
\begin{equation}
  \sum_{j:\,v_j=1} (1 - \widehat{\mathrm{FPR}}_j) \;>\; \sum_{j:\,v_j=0} (1 - \widehat{\mathrm{FNR}}_j),
  \label{eq:wmv}
\end{equation}
where $\widehat{\mathrm{FPR}}_j$ is the estimated FPR for the $j^{th}$ judge and $\widehat{\mathrm{FPR}}_j$ is the estimated FNR for the $j^{th}$ judge. Intuitively, a judge's positive vote is weighted by its estimated specificity $(1 - \mathrm{FPR})$, and its negative vote by its estimated sensitivity $(1 - \mathrm{FNR})$, so judges that are prone to false positives are down-weighted when they vote ``correct.'' Similarly, judges that are prone to false negatives are down-weighted when they vote ``incorrect.'' 

We compare four strategies for estimating $\widehat{\mathrm{FPR}}_j$ and $\widehat{\mathrm{FNR}}_j$:

\begin{enumerate}
  \item \textbf{Unweighted MV} (baseline): all weights set to 1, equivalent to simple majority vote with no error-rate estimation.
  \item \textbf{Online Bayesian}: 
  maintains a per-judge Beta posterior over FPR (and FNR), initialized with prior $\mathrm{Beta}(\alpha_0, \beta_0)$ and updated each week with a small labeled subset ($n_\ell = 8$ items),  following the conjugate Bayesian approach to annotator quality estimation~\cite{raykar2010learning}.
  The FPR estimate is the posterior mean:
  $\widehat{\mathrm{FPR}}_j = \frac{\alpha_0 + \sum \mathrm{FP}_j}{\alpha_0 + \beta_0 + \sum (\mathrm{FP}_j + \mathrm{TN}_j)}$.
  This estimator pools all labeled data globally and does not distinguish between tasks, so it can lag behind when the task mix shifts. Further details on hyperparameter selection are included in Appendix~\ref{app:experimental-details}. 
  \item \textbf{Stratified}: maintains \emph{separate} Beta posteriors per task bin (GSM8K, AIME) and aggregates using the known current mix $\pi$:
  $\widehat{\mathrm{FPR}}_j = \sum_{b} \pi_b \cdot \widehat{\mathrm{FPR}}_{j,b}$, extending annotator-difficulty models~\cite{whitehill2009whose} to the online setting with observed task categories
  This adapts to drift but requires task labels for each item.
  \item \textbf{Disagreement-based}: uses the majority vote of the \emph{other} panel members as a proxy for ground truth, inspired by the Dawid--Skene principle of estimating annotator reliability from inter-annotator agreement~\cite{dawid1979maximum}.
  For each item, if judge $j$ votes ``correct'' but the majority of other judges vote ``incorrect,'' this counts as a proxy false positive.  For each item, if judge $j$ votes ``incorrect'' but the majority of other judges vote ``correct,'' this counts as a proxy false negative. The FPR estimate is:
  $\widehat{\mathrm{FPR}}_j = \frac{\widetilde{\mathrm{FP}}_j}{\widetilde{\mathrm{FP}}_j + \widetilde{\mathrm{TN}}_j}$,
  where $\widetilde{\mathrm{FP}}$ and $\widetilde{\mathrm{TN}}$ are proxy counts derived from disagreement.
  This estimator requires \emph{no ground-truth labels and no task labels}.
\end{enumerate}

We also plot an \textbf{Oracle} line that uses the true FPR and FNR (computed on the full dataset) in Equation~\ref{eq:wmv}. This represents the best accuracy achievable by Weighted MV with perfect error-rate knowledge, and serves as an upper bound against which the online estimators can be evaluated.

\begin{figure}[t]
    \centering
    \includegraphics[width=\linewidth]{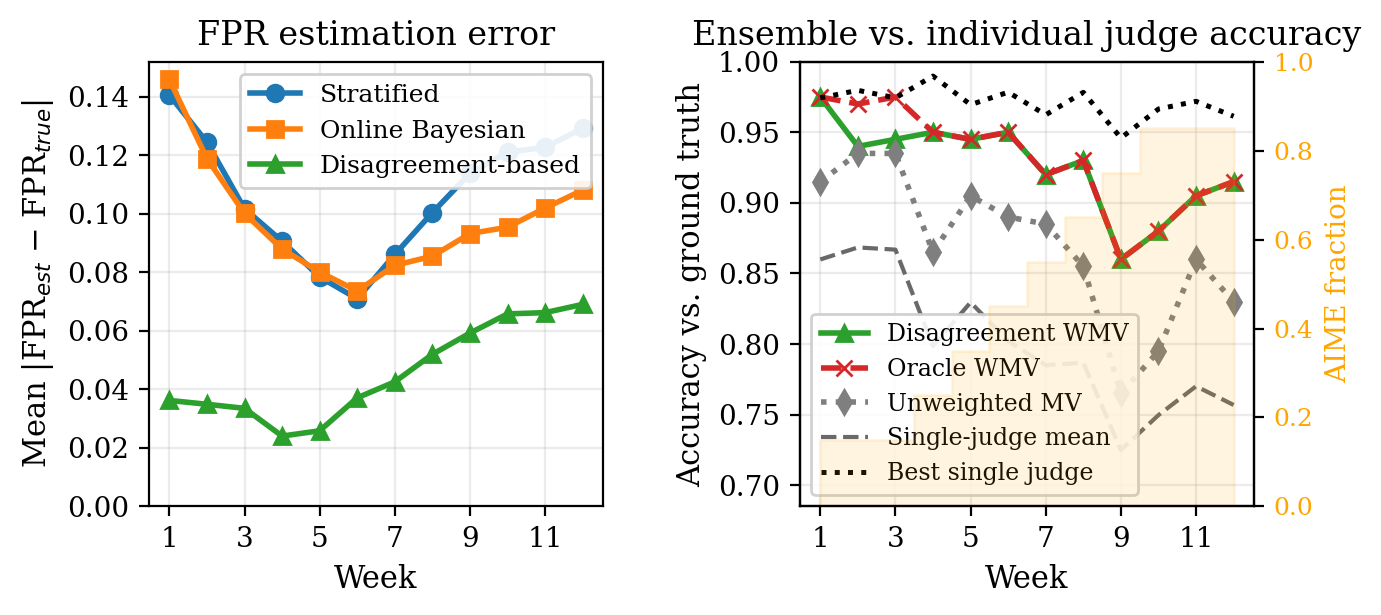}
    \caption{Performance under AIME$\leftrightarrow$GSM8K task drift over 12 weeks. \textbf{Left}: FPR estimation error (mean $|\widehat{\mathrm{FPR}}_j - \mathrm{FPR}_j|$ across judges). The disagreement-based estimator maintains error $<$0.08 throughout the drift without requiring any labels. The stratified estimator degrades as the task mix shifts away from its early calibration data. \textbf{Right}: ensemble accuracy vs.\ ground truth. All three Weighted MV variants outperform the mean single judge; the disagreement-based Weighted MV nearly matches the oracle.}
    \label{fig:drift}
\end{figure}

Figure~\ref{fig:drift} (left) shows that the disagreement-based estimator maintains consistently low FPR estimation error ($<$0.04) throughout the drift, while the online Bayesian estimator lags behind the shift and the stratified estimator degrades when the task mix diverges from its calibration data.
Figure~\ref{fig:drift} (right) shows that all Weighted MV variants outperform both the mean single-judge accuracy and the unweighted majority vote baseline.
The disagreement-based Weighted MV nearly matches the oracle, demonstrating that accurate FPR/FNR estimation is achievable purely from cross-judge disagreement patterns, without any ground-truth labels or task metadata.
\section{Ablation and Confounding Studies}
Our results above identify systematic trends in LLM judge leniency bias and show that calibrated ensembles may be a viable mitigation method. To examine whether this effect reflects a genuine bias in judging model outputs or can be explained by other variables, we conduct a set of additional experiments below.
\subsection{Test of elite bias through explicit labeling}
To look at the cause of elite model bias, we dig deeper by exploring the following question: is the elite model preference bias we observe in Section~\ref{subsec:examinee_acc_vs_bias} due to simple self recognition or something inherent to model output?

To isolate identity-driven bias from stylistic confounds, we conducted a label-swap experiment on QuALITY where identical Llama-3.3-70B responses were evaluated under three label conditions (Table~\ref{tab:label}): one without label, with labeled as response from GPT-5, and one labeled correctly. 

\begin{table}[ht]
\centering
\caption{Label-swap experiment: GPT-4.1-mini judging identical Llama-3.3-70B responses under different claimed authorship labels.}
\label{tab:label}
\small
\begin{tabular}{llrr}
\toprule
\textbf{Condition} & \textbf{Claimed source} & \textbf{Spec\%} & \textbf{FPR\%} \\
\midrule
Blind & (none) & 64.6 & 35.4 \\
Elite & GPT-5 & 72.4 & 27.6 \\
Non-elite & Llama-3.3-70B & 72.8 & 27.2 \\
\bottomrule
\end{tabular}
\end{table}

Surprisingly, both labeled conditions (elite \emph{and} non-elite) reduce FPR relative to blind: the judge becomes stricter whenever any label is present.
This suggests the label triggers more careful evaluation rather than deference, and that the primary driver of elite model preference in the blind condition is factors more complex than explicit identity recognition.

\subsection{Length Confound Analysis}
\label{sec:results:length}

A potential confound is that judges may favor \emph{longer} answers regardless of identity.
We measure length on the exact text supplied to the judge (``Final Answer: $\langle$predicted$\rangle$\textbackslash n\textbackslash nReasoning: $\langle$trace$\rangle$'') rather than the bare predicted answer.

Cohen's $d$ quantifies the practical magnitude of a group difference in units of standard deviations, making it comparable across tasks with different scales.
We define 
$$d = (\bar{x}_{\mathrm{FP}} - \bar{x}_{\mathrm{TN}}) / s_{\mathrm{pooled}},$$
where $\bar{x}_{\mathrm{FP}}$ and $\bar{x}_{\mathrm{TN}}$ are the mean answer lengths of the false-positive and true-negative groups respectively, and $s_{\mathrm{pooled}} = \sqrt{(s_{\mathrm{FP}}^2 + s_{\mathrm{TN}}^2)/2}$.
Positive $d$ indicates that wrong answers the judge \emph{incorrectly accepted} are longer than those it correctly rejected, suggesting length may have contributed to the misjudgment.
Table~\ref{tab:length} reports Cohen's $d$ for each judge--task combination.
\begin{table}[ht]
\centering
\caption{Answer-length confound on \emph{judge's} answer that includes model's reasoning trace along with final answer. Cohen's $|d| > 0.5$ is a substantive confound while $0.5 > |d| > 0.2$ is small-to-moderate confound. }
\label{tab:length}
\small
\begin{tabular}{llr}
\toprule
\textbf{Task} & \textbf{Judge} & \textbf{Cohen's $d$} \\
\midrule
AIME  & GPT-4.1-mini  & $+1.01$ \\
AIME  & DeepSeek      & $+0.33$ \\
AIME  & Llama-4-Maverick      & $+0.23$ \\
AIME  & o4-mini       & $+0.20$ \\
AIME  & GPT-5         & $-0.16$ \\
AIME  & Llama-3.3     & $+0.00$ \\
\midrule
GSM8K & DeepSeek      & $-0.25$ \\
GSM8K & Llama-4-Maverick      & $+0.11$ \\
GSM8K & o4-mini       & $-0.10$ \\
GSM8K & GPT-5         & $-0.09$ \\
GSM8K & GPT-4.1-mini  & $-0.08$ \\
GSM8K & Llama-3.3     & $-0.04$ \\
\midrule
MBPP, QuALITY & all judges & $|d| < 0.2$ \\
\bottomrule
\end{tabular}
\end{table}

Length is \emph{not} a confound on GSM8K, MBPP, or QuALITY ($|d| < 0.5$ for every judge in those tasks). 
On AIME, GPT-4.1-mini exhibits $d=+1.01$.
Inspection reveals that its 27 false positives come overwhelmingly from strong examinees (DeepSeek, GPT-5, o4-mini), whose wrong answers contain long, elaborate reasoning traces, while its 66 true negatives come primarily from weak examinees with shorter, simpler wrong answers.
GPT-4.1-mini occupies a middle capability tier: competent enough to catch obviously flawed reasoning from weak models, but unable to verify the multi-step derivations of stronger models on olympiad-level problems, so it falls back on superficial cues such as reasoning length.
Weaker judges (Llama-3.3, $d{=}0.00$; Maverick, $d{=}+0.23$) show no length effect because they accept most wrong answers indiscriminately (FPR${>}68\%$), while stronger judges (GPT-5, $d{=}{-}0.16$; o4-mini, $d{=}{+}0.20$) can verify the mathematics and reject wrong answers regardless of length.
The AIME length effect for GPT-4.1-mini is therefore best understood as a \emph{symptom} of capability-dependent bias rather than an independent confound: answer length proxies for examinee strength on hard math tasks.

\subsection{Difficulty Stratification}

Although the paper's main finding is a broader capability-dependent bias (Section~\ref{subsec:examinee_acc_vs_bias}), self-preference remains a component of that bias, and this stratification tests whether it varies with question difficulty.

For a given question $q$ and a set of $M$ examinee models, we define the difficulty:
$$k(q) = \frac{|\{m \in M : m \text{ answers } q \text{ correctly}\}|}{|M|}$$
That is, $k(q)$ is the fraction of examinee models that answer $q$ correctly, independent of any judge's evaluation. A question answered correctly by all models has $k(q) = 1$ (easy); one answered correctly by none has $k(q) = 0$ (hard). We stratify questions into three equal-width bins: hard $[0, 0.33)$, medium $[0.33, 0.67)$, and easy $[0.67, 1]$ and compute $\Delta$FPR within each. This allows us to analyze whether judge bias varies with question difficulty.

\begin{table}[t]
\centering
\caption{Difficulty-stratified $\Delta$FPR. $k(q)$ = fraction of models correct on question $q$.}
\label{tab:difficulty}
\small
\begin{tabular}{llrrr}
\toprule
& & \multicolumn{3}{c}{\textbf{$\Delta$FPR\% by $k(q)$ bin}} \\
\cmidrule(lr){3-5}
\textbf{Task} & \textbf{Metric} & \textbf{Hard} & \textbf{Medium} & \textbf{Easy} \\
\midrule
AIME & $\Delta$FPR & +18.2 & +36.2 & +15.4 \\
GSM8K & $\Delta$FPR & +16.9 & +23.7 & +31.3 \\
MBPP & $\Delta$FPR & +14.0 & +12.3 & +18.7 \\
QuALITY & $\Delta$FPR & +10.5 & +13.4 & +19.4 \\
\bottomrule
\end{tabular}
\end{table}

$\Delta$FPR, as defined in the metrics (Section~\ref{subsec:metrics}), measures the self-preference gap: the FPR when judge and examinee are the same model minus the FPR when they differ, pooled across all models within each difficulty bin.

Table~\ref{tab:difficulty}, we see that self-preference bias is present across all difficulty strata, but on GSM8K and QuALITY it \emph{increases} with easier questions ($+16.9\to+31.3$ and $+10.5\to+19.4$ respectively).
This suggests that on easy problems, where fewer models err, the judge finds the rare wrong answer from its own family more ``persuasive'' and thus harder to reject.
AIME shows a peak at medium difficulty ($+36.2$), while MBPP is relatively flat ($+12$--$19$).
The key finding is that bias persists at all difficulty levels; difficulty stratification alone does not eliminate it.

\section{Discussion and Conclusion}

We studied LLM judge reliability on absolute scoring tasks with ground-truth answers, examining both the nature of judging bias and a principled method to mitigate it.


\paragraph{Capability-dependent bias is broader than self-preference.}
Our results show that LLM judge reliability depends on the capability of both the judge and the examinee model. More accurate models generally make better judges, but judges can be accurate overall while still making asymmetric errors across examinees. We observe a pattern across all four benchmarks of capability-dependent leniency, where more capable examinee models receive more lenient judgments from \emph{all} judges. For most models, self-evaluation data points fall along the same trend as cross-model evaluations, suggesting that capability-dependent bias accounts for much of what might otherwise appear as self-preference. However, self-preference remains a distinct and meaningful effect, most prominently for GPT-5, which shows substantial self-leniency above and beyond the capability trend. We therefore view self-preference and capability-dependent bias as co-occurring effects, with the latter being broader in scope but the former remaining practically significant for frontier models. Difficulty stratification, length analysis, and label-swap experiments rule out simple confounds: the bias persists across difficulty strata, is not driven by verbosity, and is not triggered by explicit model identity labels.

\paragraph{Calibrated ensembles as a scalable remedy.}
Cross-judge majority voting already improves accuracy and reduces leniency substantially (Table~\ref{tab:mv}), but na\"{\i}ve voting treats all judges as equally reliable. Our calibrated WMV down-weights judges prone to false positives, precisely the error mode that drives capability-dependent bias. The disagreement-based estimator matches an oracle with perfect error-rate knowledge under temporal drift (Figure~\ref{fig:drift}), demonstrating that a \emph{fully unsupervised} calibration signal exists within a diverse judge panel.

\paragraph{Practical implications.}
Evaluation reports should include per-judge FPR and FNR alongside aggregate accuracy, since a judge can be accurate overall while systematically inflating the scores of stronger models. When multiple judges are available, calibrated WMV should be preferred over single-judge evaluation or unweighted majority vote. As model capabilities continue to improve, capability-dependent bias is likely to intensify, making multi-judge calibration an increasingly important safeguard.

\paragraph{Limitations and future work.}
Our model set spans three families (OpenAI, Meta, DeepSeek); broader coverage (e.g., Anthropic, Google) would strengthen generalizability. The temporal drift experiment uses real judge verdicts but simulated temporal ordering; validation on benchmarks with natural temporal structure (e.g., LiveCodeBench~\cite{jain2024livecodebench}) is an important next step. We use a single prompt template per task; prompt sensitivity may modulate bias magnitude. Finally, our disagreement-based estimator assumes the majority of panel members are approximately correct on most items; exploring its robustness when this assumption weakens (e.g., with very small or homogeneous panels) is a direction for future work.\footnote{AI writing assistants were used for editing the paper and code development. All authors take full responsibility for the content of this paper.}

\bibliographystyle{unsrt} 
\bibliography{references}

\clearpage

\appendix
\section{Additional Correlation Results}
\label{appx:correlations}
Tables~\ref{tab:corr-panel-a}--\ref{tab:corr-panel-d} report
per-evaluatee and per-judge Pearson correlation coefficients for
the four relationships plotted in Figure~\ref{fig:correlations}:
(A)~model accuracy vs.\ judge accuracy,
(B)~model accuracy vs.\ directional bias,
(C)~judge accuracy vs.\ directional bias, and
(D)~examinee accuracy vs.\ directional bias received.

\begin{table}[h]
\centering
\scriptsize
\caption{Per-evaluatee Pearson $r$ for (A) ~model accuracy vs.\ judge accuracy. Values in parentheses exclude DeepSeek-V3.2 (DS) as judge. Values where there are no additional entries in parentheses for non-DS models mean they have identical Pearson $r$ with and without the DS datapoint. Same applies in Tables~\ref{tab:corr-panel-b}--\ref{tab:corr-panel-d}.}
\label{tab:corr-panel-a}
{\fontsize{8pt}{9pt}\selectfont
\setlength{\tabcolsep}{2pt}
\begin{tabular}{lcccc}
\toprule
\textbf{Evaluatee} & \textbf{QuALITY} & \textbf{MBPP} & \textbf{GSM8K} & \textbf{AIME} \\
\midrule
DeepSeek-V3.2        & 0.97 & 0.94 & 0.96 & 0.99 \\
GPT-4.1-mini         & 0.85 (0.97) & 0.23 (0.99) & 0.62 (0.64) & 1.00 \\
GPT-5                & -0.26 (-1.00) & -0.25 (0.90) & 0.08 (0.07) & 0.12 (0.07) \\
Llama-3.3-70B        & 0.88 (0.99) & 0.46 (0.99) & 0.66 (0.65) & 0.76 (0.83) \\
Llama-4-Mav          & 0.87 (0.98) & 0.37 (0.98) & 0.86 (0.95) & 0.87 (0.92) \\
o4-mini              & 0.75 (0.78) & -0.14 (0.88) & 0.69 (0.72) & 0.51 (0.75) \\
\addlinespace
\textit{Mean (avg-level)} & 0.93 & 0.13 & 0.85 & 0.98 \\
\quad excl.\ DS judge & 0.97 & 0.95 & 0.90 & 0.99 \\
\bottomrule
\end{tabular}
}
\end{table}

\begin{table}[h]
\centering
\caption{Per-evaluatee Pearson $r$ for (B) model accuracy vs.\ directional bias. Values in parentheses exclude DeepSeek-V3.2 as judge.}
\label{tab:corr-panel-b}
{\fontsize{8pt}{9pt}\selectfont
\setlength{\tabcolsep}{2pt}
\begin{tabular}{lcccc}
\toprule
\textbf{Evaluatee} & \textbf{QuALITY} & \textbf{MBPP} & \textbf{GSM8K} & \textbf{AIME} \\
\midrule
DeepSeek-V3.2        & -0.98 & -0.90 & -0.97 & -0.99 \\
GPT-4.1-mini         & -0.62 (-0.98) & -0.80 (-0.98) & -0.70 (-0.72) & -0.96 (-0.99) \\
GPT-5                & -0.16 (-0.95) & -0.48 (-0.66) & 0.52 (0.51) & -0.86 (-0.99) \\
Llama-3.3-70B        & -0.49 (-0.98) & -0.66 (-0.95) & -0.53 & -0.01 (0.07) \\
Llama-4-Mav          & -0.56 (-0.99) & -0.74 (-1.00) & -0.91 (-1.00) & -0.48 (-0.54) \\
o4-mini              & -0.52 (-0.96) & -0.75 (-0.99) & 0.72 (0.94) & -0.97 \\
\addlinespace
\textit{Mean (avg-level)} & $-$0.71 & $-$0.76 & $-$0.81 & $-$0.89 \\
\quad excl.\ DS judge & $-$0.98 & $-$0.98 & $-$0.83 & $-$0.94 \\
\bottomrule
\end{tabular}
}
\end{table}

\begin{table}[h]
\centering
\caption{Per-evaluatee Pearson $r$ for (C) judge accuracy vs.\ directional bias. Values in parentheses exclude DeepSeek-V3.2 as judge.}
\label{tab:corr-panel-c}
{\fontsize{8pt}{9pt}\selectfont
\setlength{\tabcolsep}{2pt}
\begin{tabular}{lcccc}
\toprule
\textbf{Evaluatee} & \textbf{QuALITY} & \textbf{MBPP} & \textbf{GSM8K} & \textbf{AIME} \\
\midrule
DeepSeek-V3.2        & -1.00 & -0.98 & -1.00 & -1.00 \\
GPT-4.1-mini         & -0.92 (-1.00) & 0.37 (-1.00) & -0.94 & -0.97 (-0.99) \\
GPT-5                & 0.99 (0.94) & 0.97 (-0.29) & -0.79 (-0.80) & -0.47 (-0.24) \\
Llama-3.3-70B        & -0.83 (-1.00) & 0.34 (-0.93) & -0.91 (-0.92) & -0.52 (-0.49) \\
Llama-4-Mav          & -0.87 (-0.99) & 0.34 (-0.98) & -0.96 & -0.84 \\
o4-mini              & -0.23 (-0.93) & 0.76 (-0.86) & 0.28 (0.63) & -0.51 (-0.85) \\
\addlinespace
\textit{Mean (avg-level)} & $-$0.89 & 0.53 & $-$0.91 & $-$0.95 \\
\quad excl.\ DS judge & $-$1.00 & $-$0.96 & $-$0.91 & $-$0.98 \\
\bottomrule
\end{tabular}
}
\end{table}

\begin{table}[h]
\centering
\caption{Per-judge Pearson $r$ for (D) examinee accuracy vs.\ directional bias. Values in parentheses exclude DeepSeek-V3.2 as judge.}
\label{tab:corr-panel-d}
{\fontsize{8pt}{9pt}\selectfont
\setlength{\tabcolsep}{2pt}
\begin{tabular}{lcccc}
\toprule
\textbf{Judge} & \textbf{QuALITY} & \textbf{MBPP} & \textbf{GSM8K} & \textbf{AIME} \\
\midrule
DeepSeek-V3.2        & 0.95 & 0.93 & 0.96 & 0.30 \\
GPT-4.1-mini         & 0.99 & 0.72 & 0.67 & 0.84 \\
GPT-5                & 0.97 & 0.90 & 0.91 & -0.15 \\
Llama-3.3-70B        & 0.98 & -0.38 & 0.50 & 0.79 \\
Llama-4-Mav          & 0.99 & 0.62 & 0.85 & 0.80 \\
o4-mini              & 0.98 & 0.84 & 0.91 & -0.09 \\
\addlinespace
\textit{Mean (avg-level)} & 0.98 & 0.90 & 0.96 & 0.83 \\
\quad excl.\ DS judge & 0.99 & 0.98 & 0.96 & 0.87 \\
\bottomrule
\end{tabular}
}
\end{table}





\section{Additional Experiments}
\label{subsec:add_exp}
In addition to our main results, we conducted several sets of experiments on self-preference. Specifically, Table~\ref{tab:deltas} reports self-preference $\Delta$FPR (defined in Section~\ref{subsec:metrics} such that the FPR on others is averaged across 5 other models). 
We also estimate the 95\% bootstrap confidence intervals for the FPR rates in these experiments, which shows that $\Delta$FPR is statistically significant on most of these experiments. 
\begin{table*}[t]
\centering
\caption{Self-preference bias deltas with 95\% bootstrap CIs on FPR\textsubscript{self}. Positive $\Delta$ = self-favoring; negative = self-penalizing.}
\label{tab:deltas}
\small
\begin{tabular}{llrrrr}
\toprule
\textbf{Task} & \textbf{Judge} & \textbf{FPR\textsubscript{self}\%} & \textbf{95\% CI} & \textbf{$\Delta$FPR\%} & \textbf{$\Delta$FNR\%} \\
\midrule
\multirow{3}{*}{GSM8K}
 & Llama-3.3-70B & 29.6 & {\scriptsize[20.8, 38.8]} & $-18.2$ & $-0.6$ \\
 & GPT-4.1-mini & 28.7 & {\scriptsize[25.4, 32.0]} & $-12.2$ & $-2.9$ \\
 & GPT-5 & 82.0 & {\scriptsize[72.4, 90.7]} & $\mathbf{+67.3}$ & $-1.0$ \\
\midrule
\multirow{3}{*}{MBPP}
 & Llama-3.3-70B & 59.8 & {\scriptsize[50.5, 69.4]} & $-3.3$ & $-7.0$ \\
 & GPT-4.1-mini & 56.6 & {\scriptsize[46.0, 67.9]} & $+6.5$ & $-4.5$ \\
 & GPT-5 & 75.0 & {\scriptsize[49.9, 100.0]} & $\mathbf{+59.9}$ & $+0.4$ \\
\midrule
\multirow{3}{*}{QuALITY}
 & Llama-3.3-70B & 54.4 & {\scriptsize[50.5, 58.6]} & $-12.9$ & $+1.2$ \\
 & GPT-4.1-mini & 56.0 & {\scriptsize[52.2, 60.1]} & $-4.2$ & $-0.1$ \\
 & GPT-5 & 81.0 & {\scriptsize[76.1, 85.8]} & $\mathbf{+63.4}$ & $-3.5$ \\
\bottomrule
\end{tabular}
\end{table*}

Table~\ref{tab:harmful} isolates questions where one model is wrong and another is right and looks at how much of this subset the model incorrectly prefers itself. This marks the most harmful setting where the model is preferred despite incorrect answers. As GPT-5 had the highest rate of getting leniency, we analyze its harmful self-preference tendency.
\begin{table*}[t]
\centering
\caption{Harmful subset: FPR on \emph{own wrong answers} vs.\ \emph{others' wrong answers}. Only high-$|\Delta|$ pairs shown.}
\label{tab:harmful}
\small
\begin{tabular}{llrrr}
\toprule
\textbf{Task} & \textbf{Judge vs.\ Other} & \textbf{FPR$_\text{self}$\%} & \textbf{FPR$_\text{other}$\%} & \textbf{$\Delta$\%} \\
\midrule
GSM8K & GPT-5 vs.\ GPT-4.1-mini & 90.0 & 0.6 & $+89.4$ \\
GSM8K & GPT-5 vs.\ Llama-3.3 & 72.7 & 0.5 & $+72.2$ \\
MBPP & GPT-5 vs.\ Llama-3.3 & 100.0 & 2.0 & $+98.0$ \\
QuALITY & GPT-5 vs.\ Maverick & 85.0 & 3.1 & $+81.9$ \\
QuALITY & GPT-5 vs.\ GPT-4.1-mini & 73.8 & 6.6 & $+67.2$ \\
QuALITY & GPT-5 vs.\ Llama-3.3 & 79.3 & 3.0 & $+76.3$ \\
\midrule
GSM8K & GPT-4.1-mini vs.\ GPT-5 & 29.0 & 70.0 & $-41.0$ \\
QuALITY & GPT-4.1-mini vs.\ GPT-5 & 48.2 & 87.9 & $-39.6$ \\
\bottomrule
\end{tabular}
\end{table*}

On MBPP, GPT-5 marks its \emph{only} wrong answer as correct 100\% of the time while correctly rejecting 98.0\% of Llama's wrong answers, which indicates large asymmetry.
On QuALITY, GPT-5 forgives 85.0\% of its own errors when judging against Maverick but only 3.1\% of Maverick's errors.

GPT-4.1-mini shows the \emph{reverse} pattern when evaluating GPT-5's responses: it is more lenient on GPT-5's wrong answers than its own, hinting at elite model preference toward stronger models.

\section{Experimental Details}
\label{app:experimental-details}

\paragraph{Inference infrastructure.}
All inference is performed via managed cloud API endpoints;
no models are hosted or fine-tuned locally.
GPT-4.1-mini, GPT-5, and o4-mini are accessed through
Azure OpenAI. 
Llama-3.3-70B-Instruct, Llama-4-Maverick-17B-128E-Instruct-FP8, and
DeepSeek-V3.2 are accessed through
Azure AI Model Inference.

\paragraph{Decoding and determinism.}
We set temperature $t{=}0$ for GPT-4.1-mini,
Llama-3.3-70B, Llama-4-Maverick, and DeepSeek-V3.2 to
obtain near-deterministic output.
The OpenAI API does not support $t{=}0$ for o4-mini and GPT-5:
any value other than $t{=}1$ is rejected with an error.
This is a documented constraint for these models, which use
internal chain-of-thought tokens and do not expose a
deterministic-decoding mode.\footnote{%
  \url{https://platform.openai.com/docs/guides/reasoning}}
We do not pass a \texttt{seed} parameter.
OpenAI documents temperature-zero decoding as producing
``mostly deterministic'' results, noting that server-side
non-determinism may occasionally produce different tokens.
Each generation and scoring call is executed exactly once per
model pair; no multi-trial averaging is performed.

\paragraph{Generation parameters.}
For AIME, we set \texttt{max\_completion\_tokens} to 100{,}000
(o4-mini) and 128{,}000 (GPT-5), following
MathArena~\citep{matharena2025aime} conventions;
all other models use 16{,}384. For MBPP, we set the output-length cap to 
4096 for the reasoning models and 512 otherwise. 
For GSM8K and QuALITY, no explicit output-length cap is
set; the API default equals each model's maximum output length
(e.g.,\ 16{,}384 tokens for GPT-4.1-mini), which is never binding
because these tasks produce short responses.
All calls use a concurrency limit of 10 and retry up to 5 times
with exponential backoff (base 2\,s, ceiling 120\,s).

\paragraph{Judge input format.}
For each task, the judge receives the examinee's \emph{full response}
(not just a bare final answer) and outputs a single token:
``correct'' or ``incorrect.''
For GSM8K and AIME, the judge sees
``Final Answer: $\langle$answer$\rangle$\textbackslash n\textbackslash nReasoning:
$\langle$solution trace$\rangle$.''
For MBPP, the judge receives the submitted Python code and reasoning trace together with
the problem statement and test cases.
For QuALITY, the judge sees the full article, the question, and the
model's selected answer choice with its justification.
Length and stylistic properties of the full response are therefore
visible to the judge in all tasks, which motivates the length-confound
analysis in Section~\ref{sec:results:length}.

\paragraph{Datasets.}
Table~\ref{tab:app-datasets} summarizes the benchmarks. We do not apply any subsampling.
\begin{table}[h]
\centering
{\fontsize{7pt}{9pt}\selectfont
\setlength{\tabcolsep}{2pt}
\begin{tabular}{lllr}
\toprule
\textbf{Task} & \textbf{Source} & \textbf{Split} & \textbf{Items} \\
\midrule
GSM8K   & \texttt{openai/gsm8k}  & test  & 1{,}319 \\
AIME    & \texttt{MathArena/aime\_2025} & train\textsuperscript{\dag} & 30 \\
MBPP    & \texttt{google-research-datasets/mbpp} & test/evaluation & 500 \\
QuALITY & \texttt{tasksource/QuALITY} & dev\textsuperscript{\ddag} & 2{,}202 \\
\bottomrule
\end{tabular}
\caption{Datasets.
\textsuperscript{\dag}Only available split.
\textsuperscript{\ddag}Items missing required fields removed during
preprocessing;
test-split labels are not publicly released.}
\label{tab:app-datasets}
}
\end{table}
\paragraph{Data completeness.}
For GSM8K, AIME, and MBPP, every judge--examinee cell is 100\% complete.
For QuALITY, items where answer generation failed after exhausting
retry attempts are excluded from scoring; the evaluated item count
ranges from 1{,}638 to 2{,}287 per judge--examinee pair, depending on
the target model's generation success rate. Most of these failures are
transient (rate-limiting and timeouts) and not content-dependent, while very few failures occur due to model safety guardrails. 

\paragraph{Statistical analysis.}
All confidence intervals use $B = 1000$ bootstrap resamples
(seed~$= 42$) with the percentile method at the 95\% level.

\paragraph{WMV simulation hyperparameters.}
\label{par:wmv-hps}
The simulation (Section~\ref{sec:mitigation}) samples $n = 40$ items
per week for 12 weeks, with a labelled subset of $n_\ell = 8$ items
per week used to update the Bayesian and Stratified estimators.
A minimum of 5 judges per item is required.
The simulation seed is 0.

The \textbf{Online Bayesian} estimator (item~2 in
Section~\ref{sec:mitigation}) initializes a Beta prior for each
judge's FPR and FNR.
For FPR, the prior is $\mathrm{Beta}(\alpha_0, \beta_0)$ with
$\alpha_0 = 0.4 \times 20 = 8$ and $\beta_0 = 0.6 \times 20 = 12$
(prior mean $= 0.4$).
For FNR, $\alpha_0 = 0.2 \times 20 = 4$, $\beta_0 = 0.8 \times 20 = 16$
(prior mean $= 0.2$).
The total pseudo-count of 20 means the prior is equivalent to
having observed 20 labelled items, and is therefore dominated by
real observations after $\approx$3 labelling rounds ($3 \times 8 = 24$
observed items).
The prior means (0.4 for FPR, 0.2 for FNR) are set above empirical
averages so that the posterior can update downward toward the data;
this avoids underweighting good judges during early weeks.

The \textbf{Stratified} estimator (item~3) maintains separate priors
per task bin with half the pseudo-count (strength~$=10$), yielding
$\mathrm{Beta}(4, 6)$ for FPR and $\mathrm{Beta}(2, 8)$ for FNR in
each bin. The lower strength allows faster adaptation within each
stratum.

The \textbf{Disagreement-based} estimator (item~4) uses no priors;
it derives FPR and FNR counts purely from inter-judge disagreement.

We swept prior means over $\{0.1, 0.2, 0.3, 0.4, 0.5\}$ and pseudo-count strengths over $\{2, 10, 40\}$ (symmetric, 30 configurations), and additionally swept $(p_{\mathrm{FPR}}, p_{\mathrm{FNR}}) \in \{0.05, 0.15, 0.25, 0.35, 0.50\}^2$ at strength~2 (asymmetric, 50 configurations). All 80 tested configurations outperform unweighted majority vote (90.8--95.0\% vs.\ 87.0\%); accuracy varies by ${<}3$~pp across the grid. We selected $p_{\mathrm{FPR}}{=}0.4$, $p_{\mathrm{FNR}}{=}0.2$, strength${=}10$ as a moderately informative prior that reflects the empirical $\mathrm{FPR} \gg \mathrm{FNR}$ asymmetry observed across judges while converging to data-driven estimates within 2--3 labelling rounds (${\approx}16$--$24$ observations per judge).

In the WMV formula (Equation~\ref{eq:wmv}), each judge's positive-vote
weight is $(1 - \widehat{\mathrm{FPR}}_j)$ and each negative-vote
weight is $(1 - \widehat{\mathrm{FNR}}_j)$.
A \emph{weight floor} $\varepsilon = 0.02$ is applied:
weights are clipped to $\max(\varepsilon,\, 1 - \widehat{\mathrm{FPR}}_j)$
and $\max(\varepsilon,\, 1 - \widehat{\mathrm{FNR}}_j)$,
ensuring no judge is fully silenced even if its estimated error rate
approaches~1.

\end{document}